\documentclass[runningheads,a4paper]{llncs}
\usepackage{amsmath}
\usepackage{amssymb}

\usepackage{graphicx}

\usepackage{cite}
\usepackage{nohyperref}  % This makes hyperref commands do nothing without errors
\usepackage{url}  % This makes \url work

\usepackage{color} 
\usepackage{epsfig}
\usepackage{amsfonts}
\usepackage[linesnumbered,vlined,ruled]{algorithm2e}

\def\squareforqed{\hbox{\rlap{$\sqcap$}$\sqcup$}}
\def\qed{\ifmmode\squareforqed\else{\unskip\nobreak\hfil
\penalty50\hskip1em\null\nobreak\hfil\squareforqed
\parfillskip=0pt\finalhyphendemerits=0\endgraf}\fi}
\newcommand{\keywords}[1]{\par\addvspace\baselineskip
\noindent\keywordname\enspace\ignorespaces#1}

\begin{document}

\mainmatter  % start of an individual contribution

% 12 pages maximum for ICONIP 2018!!!

\title{Continuous Trade-off Optimization between Fast and Accurate Deep Face Detectors\vspace{-0.4cm}}

\titlerunning{Trade-off Optimization between Fast and Accurate Deep Face Detectors}

\author{\vspace{-0.2cm}Petru Soviany\inst{1}
			\and
        	Radu Tudor Ionescu\inst{1,2}
}

\institute{University of Bucharest, 14 Academiei, Bucharest, Romania,\\
\and
SecurifAI, 24 Mircea Vod\u{a}, Bucharest, Romania\\
\email{petru.soviany@yahoo.com, raducu.ionescu@gmail.com}
}

\date{\today}

\toctitle{Continuous Trade-off Optimization between Fast and Accurate Deep Face Detectors}
\tocauthor{Petru Soviany, Radu Tudor Ionescu}
\maketitle

%%%%%%%%% ABSTRACT
\begin{abstract}
\vspace{-0,4cm}
Although deep neural networks offer better face detection results than shallow or handcrafted models, their complex architectures come with higher computational requirements and slower inference speeds than shallow neural networks. In this context, we study five straightforward approaches to achieve an optimal trade-off between accuracy and speed in face detection. All the approaches are based on separating the test images in two batches, an \emph{easy} batch that is fed to a faster face detector and a \emph{difficult} batch that is fed to a more accurate yet slower detector.
%The difference between the five approaches is the criterion used for splitting the images in two batches. The criteria are the class-agnostic image difficulty score (easier images go into the easy batch), the person-aware image difficulty score (easier images go into the easy batch), the number of detected faces (images with less faces go into the easy batch), the average size of the detected faces (images with bigger faces go into the easy batch), and the number of detected faces divided by their average size (images with less and bigger faces go into the easy batch). The first two approaches are based on an image difficulty predictor, while the other three approaches employ a faster single-shot detector (SSD) to estimate the number of faces and each of their sizes. 
We conduct experiments on the AFW and the FDDB data sets, using MobileNet-SSD as the fast face detector and S$^3$FD (Single Shot Scale-invariant Face Detector) as the accurate face detector, both models being pre-trained on the WIDER FACE data set. Our experiments show that the proposed difficulty metrics compare favorably to a random split of the images. %Moreover, we only lose about $1$-$2\%$ in accuracy (yet gaining more than $3$-$5\%$ over the random split), while slashing nearly half of the inference time. 
\vspace{-0,3cm}
\end{abstract}

%%%%%%%%% BODY TEXT
\vspace{-0,9cm}
\section{Introduction}
\vspace*{-0.2cm}

Face detection, the task of predicting where faces are located in an image, is one of the most well-studied problems in computer vision, since it represents a prerequisite for many other tasks such as face recognition~\cite{Parkhi-BMVC-2015}, facial expression recognition~\cite{radu-WREPL-2013,Georgescu-arxiv-2018}, age estimation, gender classification and so on. Inspired by the recent advances in deep object detection~\cite{Ren-NIPS-2015,He-ICCV-2017}, researchers have proposed very deep neural networks~\cite{Chen-ECCV-2016,Jiang-FG-2017,Qin-CVPR-2016,Yang-ICCV-2015,Zhang-ICCV-2017} as a solution to the face detection task, providing significant accuracy improvements. Although deep models~\cite{He-CVPR-2016,He-ICCV-2017} generally offer better results than shallow~\cite{Howard-arXiv-2017,Hinton-NIPS-2012} or handcrafted models~\cite{Viola-2004,Felzenszwalb-2010}, their complex architectures come with more computational requirements and slower inference speeds. As an alternative for environments with limited resources, e.g. mobile devices, researchers have proposed shallower neural networks~\cite{Howard-arXiv-2017} that provide fast but less accurate results. In this context, we believe it is relevant to propose and evaluate an approach that allows to set the trade-off between accuracy and speed in face detection on a continuous scale. Based on the same principles described in~\cite{Soviany-SYNASC-2018}, we hypothesize that using more complex and accurate face detectors for \emph{difficult} images and less complex and fast face detectors for \emph{easy} images will provide an optimal trade-off between accuracy and speed, without ever having to change anything about the face detectors. The only problem that prevents us from testing our hypothesis in practice is finding an approach to classify the images into easy or hard. In order to be useful in practice, the approach also has to work fast enough, e.g. at least as fast as the faster face detector. To this end, we propose and evaluate five simple and straightforward approaches to achieve an optimal trade-off between accuracy and speed in face detection. All the approaches are based on separating the test images in two batches, an \emph{easy} batch that is fed to the faster face detector and a \emph{hard} (or \emph{difficult}) batch that is fed to the more accurate face detector. The difference between the five approaches is the criterion used for splitting the images in two batches. The first approach assigns a test image to the easy or the hard batch based on the class-agnostic image difficulty score, which is estimated using a recent approach for image difficulty prediction introduced by Ionescu et al.~\cite{Ionescu-CVPR-2016}. The image difficulty predictor is obtained by training a deep neural network to regress on the difficulty scores produced by human annotators. The second approach is based on a person-aware image difficulty predictor, which is trained only on images containing the class \emph{person}. The other three approaches used for splitting the test images (into easy or hard) employ a faster single-shot face detector, namely MobileNet-SSD~\cite{Howard-arXiv-2017}, in order estimate the number of faces and each of their sizes. The third and the fourth approaches independently consider the number of detected faces (images with less faces go into the easy batch) and the average size of the faces (images with bigger faces go into the easy batch), while the fifth approach is based on the number of faces divided by their average size (images with less and bigger faces go into the easy batch). If one of the latter three approaches classifies an image as easy, there is nothing left to do (we can directly return the detections provided by MobileNet-SSD). %, unless we want to use a different (slower, but more accurate) face detector.
% In order to achieve a trade-off between accuracy and speed in object detection, we apply the image difficulty predictor on the test images to split them into easy versus hard (difficult) images. Once separated, the easy images are sent to the faster single-stage detector, while the hard images are sent to the more accurate two-stage detector. 
Our experiments on the AFW~\cite{Zhu-CVPR-2012} and the FDDB~\cite{Jain-UMAS-2010} data sets show that using the class-agnostic or person-aware image difficulty as a primary cue for splitting the test images compares favorably to a random split of the images. However, the other three approaches, which are frustratingly easy to implement, can also produce good results. Among the five proposed methods, the best results are obtained by the class-agnostic image difficulty predictor. This approach shortens the processing time nearly by half, while reducing the Average Precision of the Single Shot Scale-invariant Face Detector (S$^3$FD)~\cite{Zhang-ICCV-2017} from $0.9967$ to no less than $0.9818$ on AFW. Moreover, all our methods are simple and have the advantage that they allow us to choose the desired trade-off on a continuous scale.

The rest of this paper is organized as follows. Recent related works on face detection are presented in Section~\ref{sec_RelatedWork}. Our methodology is described in Section~\ref{sec_Method}. The face detection experiments are presented in Section~\ref{sec_Experiments}. Finally, we draw our conclusions in Section~\ref{sec_Conclusion}.

\vspace*{-0.3cm}
\section{Related Work}
\label{sec_RelatedWork}
\vspace*{-0.3cm}

To our knowledge, there are no previous works that study the trade-off between accuracy and speed for deep face detection. However, there are works~\cite{Huang-CVPR-2017,Soviany-SYNASC-2018} that study the trade-off between accuracy and speed for the more general task of object detection. Huang et al.~\cite{Huang-CVPR-2017} have tested different configurations of deep object detection frameworks by changing various components and parameters in order to find optimal configurations for specific scenarios, e.g. deployment on mobile devices. Different from their approach, Soviany et al.~\cite{Soviany-SYNASC-2018} treat the various object detection frameworks as black boxes. Instead of looking for certain configurations, they propose a framework that allows to set the trade-off between accuracy and speed on a continuous scale, by specifying the point of splitting the test images into easy versus hard, as desired. We build our work on top of the work of Soviany et al.~\cite{Soviany-SYNASC-2018}, by considering various splitting strategies for a slightly different task: face detection.

In the rest of this section, we provide a brief description of some of the most recent deep face detectors, in chronological order. CascadeCNN~\cite{Li-CVPR-15} is one of the first models to successfully use convolutional neural networks (CNN) for face detection. Its cascading architecture is made of three CNNs for face versus non-face classification and another three CNNs for bounding box calibration. At each step in the cascade setup, a number of detections are dropped, while the others are passed to the next CNN. By changing the thresholds required in this process, a certain trade-off between accuracy and speed can be obtained.  
Jian et al.~\cite{Jiang-FG-2017} use Faster R-CNN to detect faces. Faster R-CNN~\cite{Ren-NIPS-2015} is a very accurate region-based deep detection model which improves Fast R-CNN~\cite{Girshick-ICCV-2015} by introducing the Region Proposal Networks (RPN). It uses a fully convolutional network that can predict object bounds at every location in order to solve the challenge of selecting the right regions. In the second stage, the regions proposed by the RPN are used as an input for the Fast R-CNN model, which will provide the final object detection results.
S$^3$FD~\cite{Zhang-ICCV-2017} is a highly accurate real-time face detector, based on the anchor model used initially for object detection~\cite{Liu-ECCV-2016,Ren-NIPS-2015}. In order to solve the limitations on small objects (faces) of these methods, S$^3$FD introduces a scale compensation anchor matching strategy to improve recall, and a max-out background label to reduce the false positive detections. In order to handle different scales of faces, it uses a scale-equitable face detection framework, tiling anchors on a wide range of layers, while also designing many different anchor scales.
MobileNets~\cite{Howard-arXiv-2017} are a set of lightweight models that can be used for classification, detection and segmentation tasks. They are built on depth-wise separable convolutions with a total of $28$ layers and can be further parameterized, making them very suitable for mobile devices. The fast speeds and the low computational requirements of MobileNets make up for the fact that they do not achieve the accuracy of the very-deep models. The experimental results show they can also be successfully used for face detection.
% In this paper. we tried to find the best trade-off between these models, in order to improve the time required for the detection task, while keeping the accuracy as high as possible.

\vspace*{-0.3cm}
\section{Methodology}
\label{sec_Method}
\vspace*{-0.3cm}

Humans learn much better when the examples are not randomly presented, but organized in a meaningful order which gradually illustrates more complex concepts. %This is essentially reflected in all the curricula taught in schooling systems around the world. 
Bengio et al.~\cite{Bengio-ICML-2009} have explored easy-to-hard strategies to train machine learning models, showing that machines can also benefit from learning by gradually adding more difficult examples. They introduced a general formulation of the easy-to-hard training strategies known as \emph{curriculum learning}. However, we can hypothesize that an \emph{easy-versus-hard} strategy can also be applied at test time in order to obtain an optimal trade-off between accuracy and processing speed. For example, if we have two types of machines (one that is simple and fast but less accurate, and one that is complex and slow but more accurate), we can devise a strategy in which the fast machine is fed with the easy test samples and the complex machine is fed with the difficult test samples. This kind of strategy will work as desired especially when the fast machine can reach an accuracy level that is close to the accuracy level of the complex machine for the easy test samples. Thus, the complex and slow machine will be used only when it really matters, i.e. when the examples are too difficult for the fast machine. The only question that remains is how to determine if an example is easy or hard in the first place. If we focus our interest on image data, the answer to this question is provided by the recent work of Ionescu et al.~\cite{Ionescu-CVPR-2016}, which shows that the difficulty level of an image (with respect to a visual search task) can be automatically predicted. With an image difficulty predictor at our disposal, we have a first way to test our hypothesis in the context of face detection from images. However, if we further focus our interest on the specific task of face detection in images, we can devise additional criteria for splitting the images into easy or hard. One criterion is to consider an image difficulty predictor that is specifically trained on images with people, i.e. a person-aware image difficulty predictor. Other criteria can be developed by considering the output of a very fast single-shot face detector, e.g. MobileNet-SSD~\cite{Howard-arXiv-2017}. These criteria are the number of detected faces in the image, the average size of the detected faces, and the number of detected faces divided by their average size.

\begin{algorithm}[!tpb]
%\small{
\caption{Easy-versus-Hard Face Detection\label{alg_easy_to_hard}}

\textbf{Input}: 

$I$ -- an input test image;

$D_{fast}$ -- a fast but less accurate face detector;

$D_{slow}$ -- a slow but more accurate face detector;

$C$ -- a criterion function used for dividing the images;

$t$ -- a threshold for dividing images into easy or hard;

\BlankLine
\textbf{Computation}:

\If{$C(I) \leq t$}
{
	$B \leftarrow D_{fast}(I)$\;
}
\Else
{
	$B \leftarrow D_{slow}(I)$\;
}

\BlankLine
\textbf{Output}: 

$B$ -- the set of predicted bounding boxes.
%}
\end{algorithm}

To obtain an optimal trade-off between accuracy and speed in face detection, we propose to employ a more complex face detector, e.g. S$^3$FD~\cite{Zhang-ICCV-2017}, for difficult test images and a less complex face detector, e.g. MobileNet-SSD~\cite{Howard-arXiv-2017}, for easy test images. Our simple easy-versus-hard strategy is formally described in Algorithm~\ref{alg_easy_to_hard}. Since we apply this strategy at test time, the face detectors as well as the image difficulty predictors can be independently trained beforehand. This allows us to directly apply state-of-the-art pre-trained face detectors~\cite{Howard-arXiv-2017,Zhang-ICCV-2017}, essentially as black boxes. It is important to note that we use one of the following five options as the criterion function $C$ in Algorithm~\ref{alg_easy_to_hard}:
\begin{enumerate}
\item a class-agnostic image difficulty predictor that estimates the difficulty of the input image;
\item a person-aware image difficulty predictor that estimates the difficulty of the input image;
\item a fast face detector that returns the number of faces detected in the input image (\emph{less faces} is easier);
\item a fast face detector that returns the average size of the faces detected in the input image (\emph{bigger faces} is easier);
\item a fast face detector that returns the number of detected faces divided by their average size (\emph{less and bigger faces} is easier).
\end{enumerate}
We note that if either one of the last three criteria are employed in Algorithm~\ref{alg_easy_to_hard}, and if the fast face detector used in the criterion function $C$ is the same as $D_{fast}$, we can slightly optimize Algorithm~\ref{alg_easy_to_hard} by applying the fast face detector only once, when the input image $I$ turns out to be easy. Another important note is that, for the last three criteria, we consider an image to be \emph{difficult} if the fast detector does not detect any face.
Our algorithm has only one parameter, namely the threshold $t$ used for dividing images into easy or hard. This parameter depends on the criterion function and it needs to be tuned on a validation set in order to achieve a desired trade-off between accuracy and time. While the last three splitting criteria are frustratingly easy to implement when a fast pre-trained face detector is available, we have to train our own image difficulty predictors as described below.

\noindent
{\bf Image difficulty predictors.}
We build our image difficulty prediction models based on CNN features and linear regression with $\nu$-Support Vector Regression ($\nu$-SVR)~\cite{Chang-NC-2002}. For a faster processing time, we consider a rather shallow pre-trained CNN architecture, namely VGG-f~\cite{Chatfield-BMVC-14}. The CNN model is trained on the
% ImageNet Large-Scale Visual Recognition Challenge 
ILSVRC benchmark~\cite{Russakovsky2015}.
We remove the last layer of the CNN model and use it to extract deep features from the fully-connected layer known as \emph{fc7}. The $4096$ CNN features extracted from each image are normalized using the $L_2$-norm. The normalized feature vectors are then used to train a $\nu$-SVR model to regress to the ground-truth difficulty scores provided by Ionescu et al.~\cite{Ionescu-CVPR-2016} for the PASCAL VOC 2012 data set~\cite{Pascal-VOC-2012}. We use the learned model as a continuous measure to automatically predict image difficulty. We note that Ionescu et al.~\cite{Ionescu-CVPR-2016} showed that the resulted image difficulty predictor is class-agnostic. Since our focus is on face detection, it is perhaps more useful to consider a class-specific image difficulty predictor. For this reason we train a different image difficulty predictor by selecting only the PASCAL VOC 2012 images that contain the class \emph{person}. As these images are likely to contain faces, the person-aware difficulty predictor could be more appropriate for the task at hand.
% Our predictor attains a Kendall's $\tau$ correlation coefficient~\cite{upton-dict-stat-2008} of $0.441$ on the test set of Ionescu et al.~\cite{Ionescu-CVPR-2016}. We note that Ionescu et al.~\cite{Ionescu-CVPR-2016} obtain a higher Kendall's $\tau$ score ($0.472$) using a deeper CNN architecture~\cite{Simonyan-ICLR-14} along with VGG-f. 
Both image difficulty predictors are based on the VGG-f architecture, which is faster than the considered face detectors, including MobileNet-SSD~\cite{Howard-arXiv-2017}, and it reduces the computational overhead at test time.

\vspace*{-0.2cm}
\section{Experiments}
\label{sec_Experiments}
\vspace*{-0.2cm}

\subsection{Data Sets}
\vspace*{-0.1cm}

We perform face detection experiments on the AFW~\cite{Zhu-CVPR-2012} and the FDDB~\cite{Jain-UMAS-2010} data sets. The AFW data set consists of $205$ images with $473$ labeled faces, while the FDDB data set consists of $2845$ images that contain $5171$ face instances.

\vspace*{-0.2cm}
\subsection{Evaluation Details}
\vspace*{-0.1cm}

\noindent
{\bf Evaluation Measures.}
On the FDDB data set, the performance of the face detectors is commonly evaluated using the area under the discrete ROC curve (DiscROC) or the area under the continuous ROC curve (ContROC), as defined by Jain et al.~\cite{Jain-UMAS-2010}. 
On the other hand, the performance of the face detectors on the AFW data set is typically evaluated using the Average Precision (AP) metric, which is based on the ranking of detection scores~\cite{Pascal-VOC-2010}. The Average Precision is given by the area under the precision-recall (PR) curve for the detected faces. The PR curve is constructed by mapping each detected bounding box to the most-overlapping ground-truth bounding box, according to the Intersection over Union (IoU) measure, but only if the IoU is higher than $50\%$~\cite{Everingham-IJCV-2015}. %Then, the detections are sorted in decreasing order of their scores. Precision and recall values are computed each time a new positive sample is recalled. The PR curve is given by plotting the precision and recall pairs as lower scored detections are progressively included.

%\vspace*{-0.1cm}
%\subsection{Models and Baselines}
%\vspace*{-0.1cm}
\noindent
{\bf Models and Baselines.}
We use S$^3$FD~\cite{Zhang-ICCV-2017} as our accurate model for predicting bounding boxes, and experiment with the pre-trained version available at {https://github.com/sfzhang15/SFD}. As our fast detector, we choose the pre-trained version of MobileNet-SSD from {https://github.com/yeephycho/tensorflow-face-detection}, slightly modified. For both models, which are pre-trained on the WIDER FACE data set~\cite{Yang-CVPR-2016}, we set the confidence threshold to $0.5$.

The main goal of the experiments is to compare our five different strategies for splitting the images between the fast detector (MobileNet-SSD) and the accurate detector (S$^3$FD) with a baseline strategy that splits the images randomly. To reduce the accuracy variation introduced by the random selection of the baseline strategy, we repeat the experiments for $5$ times and average the resulted scores. We note that all standard deviations are lower than $0.5\%$. We consider various splitting points starting with a $100\%-0\%$ split (equivalent with applying the fast MobileNet-SSD only), going through three intermediate splits ($75\%-25\%$, $50\%-50\%$, $25\%-75\%$) and ending with a $0\%-100\%$ split (equivalent with applying the accurate S$^3$FD only). % To obtain these splitting points, we individually tune the parameter $t$ of Algorithm~\ref{alg_easy_to_hard} on the PASCAL VOC 2007 validation set, for each splitting criterion.

%The second strategy is based on splitting the images into easy or hard, according to the difficulty scores assigned by our image difficulty predictor, as described in Section~\ref{sec_Method}.

\vspace*{-0.2cm}
\subsection{Results and Discussion}
\vspace*{-0.1cm}

\begin{table*}[!tpb]
\small{
\caption{Average Precision (AP) and time comparison between MobileNet-SSD~\cite{Howard-arXiv-2017}, S$^3$FD~\cite{Zhang-ICCV-2017} and various combinations of the two face detectors on AFW. The test data is partitioned based on a random split (baseline) or five easy-versus-hard splits given by: the class-agnostic image difficulty score, the person-aware image difficulty score, the number of faces ($n$), the average size of the faces ($avg$), and the number of faces divided by their average size ($n / avg$). For the random split, we report the AP over 5 runs to reduce bias. The reported times are measured on a computer with Intel Core i7 $2.5$ GHz CPU and $16$ GB of RAM.}
\vspace*{-0.1cm}
\begin{center}
\begin{tabular}{|l|c|c|c|c|c|}
\hline
                                        & \multicolumn{5}{|c|}{MobileNet-SSD (left) to S$^3$FD (right)}\\
\cline{2-6}
                                        & $100\%\!-\!0\%$   & $75\%\!-\!25\%$   & $50\%\!-\!50\%$   & $25\%\!-\!75\%$   & $0\%\!-\!100\%$ \\
\hline
\hline
Splitting Criterion & \multicolumn{5}{|c|}{Average Precision (AP)}\\
\hline
$(1)$ Random (baseline)	     	        & $0.8910$		& $0.9116$		& $0.9355$		& $0.9640$		& $0.9967$\\
$(2)$ Class-agnostic difficulty	        & $0.8910$		& $0.9327$		& $0.9818$		& $0.9923$		& $0.9967$\\
$(3)$ Person-aware difficulty	        & $0.8910$		& $0.9268$		& $0.9804$		& $0.9900$		& $0.9967$\\
$(4)$ Number of faces ($n$)             & $0.8910$		& $0.9591$		& $0.9741$		& $0.9912$		& $0.9967$\\
$(5)$ Average face size ($avg$)         & $0.8910$		& $0.9250$		& $0.9565$		& $0.9747$		& $0.9967$\\
$(6)$ $n / avg$                         & $0.8910$		& $0.9571$		& $0.9776$		& $0.9873$		& $0.9967$\\
\hline
\hline
Component           & \multicolumn{5}{|c|}{Time (seconds)}\\
\hline
$(2,3)$ Image difficulty                & -             & $0.05$        & $0.05$        & $0.05$        & - \\
$(4,5,6)$ Estimation of $n$, $avg$      & -		        & $0.28$		& $0.28$		& $0.28$		& - \\
Face detection                          & $0.28$		& $0.68$		& $1.08$		& $1.49$		& $1.89$\\
Face detection + $(2,3)$                & $0.28$		& $0.73$		& $1.13$		& $1.54$		& $1.89$\\
Face detection + $(4,5,6)$              & $0.28$		& $0.75$		& $1.22$		& $1.70$		& $1.89$\\
\hline
\end{tabular}
\end{center}
\label{Tab_AFW_Results}
}
\vspace*{-0.8cm}
\end{table*}

Table~\ref{Tab_AFW_Results} presents the AP scores and the processing times of MobileNet-SSD~\cite{Howard-arXiv-2017}, S$^3$FD~\cite{Zhang-ICCV-2017} and several combinations of the two face detectors, on the AFW data set. Different model combinations are obtained by varying the percentage of images processed by each detector. The table includes results starting with a $100\%-0\%$ split (equivalent with MobileNet-SSD~\cite{Howard-arXiv-2017} only), going through three intermediate splits ($75\%-25\%$, $50\%-50\%$, $25\%-75\%$) and ending with a $0\%-100\%$ split (equivalent with S$^3$FD~\cite{Zhang-ICCV-2017} only). In the same manner, Table~\ref{Tab_FDDB_Results} shows the results for the same combinations of face detectors on the FDDB data set. While the results of various model combinations are listed on different columns in Table~\ref{Tab_AFW_Results} and Table~\ref{Tab_FDDB_Results}, the results of various splitting strategies are listed on separate rows.

We first analyze the detection accuracy and the processing time of the two individual face detectors, namely MobileNet-SSD~\cite{Howard-arXiv-2017} and S$^3$FD~\cite{Zhang-ICCV-2017}. On AFW, S$^3$FD reaches an AP score of $0.9967$ in about $1.89$ seconds per image, while on FDDB, it reaches a DistROC score of $0.9750$ in about $1.17$ seconds per image. MobileNet-SDD is more than four times faster, attaining an AP score of $0.8910$ on AFW and a DistROC score of $0.8487$ on FDDB, in just $0.28$ seconds per image.
We next analyze the average face detection times per image of the various model combinations on AFW. As expected, the time improves by about $19\%$ when running MobileNet-SSD on $25\%$ of the test set and S$^3$FD on the rest of $75\%$. On the $50\%-50\%$ split, the processing time is nearly $40\%$ shorter than the time required for processing the entire test set with S$^3$FD only ($0\%-100\%$ split). On the $75\%-25\%$ split, the processing time further improves by $63\%$. As the average time per image of S$^3$FD is shorter on FDDB, the time improvements are close, but not as high. The improvements in terms of time are $15\%$ for the $25\%-75\%$ split, $34\%$ for the $50\%-50\%$ split, and $55\%$ for the $75\%-25\%$ split. We note that unlike the random splitting strategy, the easy-versus-hard splitting strategies require additional processing time, either for computing the difficulty scores or for estimating the number of faces and their average size. The image difficulty predictors run in about $0.05$ seconds per image, while the MobileNet-SSD detector (used for estimating the number of faces and their average size) runs in about $0.28$ seconds per image. Hence, the extra time required by the two splitting strategies based on image difficulty is almost insignificant with respect to the total time required by the various combinations of face detectors. For instance, in the $50\%-50\%$ split with MobileNet-SSD and S$^3$FD, the difficulty predictors account for roughly $4\%$ of the total processing time ($0.05$ out of $1.13$ seconds per image) for an image taken from AFW.

\begin{table*}[!t]
\small{
\caption{Area under the discrete and the continuous ROC curves and time comparison between MobileNet-SSD~\cite{Howard-arXiv-2017}, S$^3$FD~\cite{Zhang-ICCV-2017} and various combinations of the two face detectors on FDDB. The test data is partitioned based on a random split (baseline) or five easy-versus-hard splits given by: the class-agnostic image difficulty score, the person-aware image difficulty score, the number of faces ($n$), the average size of the faces ($avg$), and the number of faces divided by their average size ($n / avg$). For the random split, we report the scores over 5 runs to reduce bias. The reported times are measured on a computer with Intel Core i7 $2.5$ GHz CPU and $16$ GB of RAM.}
\vspace*{-0.1cm}
\begin{center}
\begin{tabular}{|l|c|c|c|c|c|}
\hline
                                        & \multicolumn{5}{|c|}{MobileNet-SSD (left) to S$^3$FD (right)}\\
\cline{2-6}
                                        & $100\%\!-\!0\%$   & $75\%\!-\!25\%$   & $50\%\!-\!50\%$   & $25\%\!-\!75\%$   & $0\%\!-\!100\%$ \\
\hline
\hline
Splitting Criterion & \multicolumn{5}{|c|}{Area under the discrete ROC curve (DiscROC)}\\
\hline
$(1)$ Random (baseline)	     	        & $0.8487$		& $0.8775$		& $0.9110$		& $0.9432$		& $0.9750$\\
$(2)$ Class-agnostic difficulty	        & $0.8487$		& $0.9131$		& $0.9493$		& $0.9673$		& $0.9750$\\
$(3)$ Person-aware difficulty	        & $0.8487$		& $0.9139$		& $0.9441$		& $0.9659$		& $0.9750$\\
$(4)$ Number of faces ($n$)             & $0.8487$		& $0.9102$		& $0.9379$		& $0.9551$		& $0.9750$\\
$(5)$ Average face size ($avg$)         & $0.8487$		& $0.9187$		& $0.9475$		& $0.9638$		& $0.9750$\\
$(6)$ $n / avg$                         & $0.8487$		& $0.9214$		& $0.9493$		& $0.9626$		& $0.9750$\\
\hline
                     & \multicolumn{5}{|c|}{Area under the continuous ROC curve (ContROC)}\\
\hline
$(1)$ Random (baseline)	     	        & $0.7092$		& $0.7425$		& $0.7756$		& $0.8096$		& $0.8432$\\
$(2)$ Class-agnostic difficulty	        & $0.7092$		& $0.7729$		& $0.8096$		& $0.8302$		& $0.8432$\\
$(3)$ Person-aware difficulty	        & $0.7092$		& $0.7737$		& $0.8052$		& $0.8291$		& $0.8432$\\
$(4)$ Number of faces ($n$)             & $0.7092$		& $0.7741$		& $0.8032$		& $0.8221$		& $0.8432$\\
$(5)$ Average face size ($avg$)         & $0.7092$		& $0.7776$		& $0.8082$		& $0.8281$		& $0.8432$\\
$(6)$ $n / avg$                         & $0.7092$		& $0.7824$		& $0.8111$		& $0.8275$		& $0.8432$\\
\hline
\hline
Component           & \multicolumn{5}{|c|}{Time (seconds)}\\
\hline
$(2,3)$ Image difficulty                & -             & $0.05$        & $0.05$        & $0.05$        & - \\
$(4,5,6)$ Estimation of $n$, $avg$      & -		        & $0.27$		& $0.27$		& $0.27$		& - \\
Face detection                          & $0.27$		& $0.51$		& $0.73$		& $0.94$		& $1.17$\\
Face detection + $(2,3)$                & $0.27$		& $0.56$		& $0.78$		& $0.99$		& $1.17$\\
Face detection + $(4,5,6)$              & $0.27$		& $0.58$		& $0.86$		& $1.14$		& $1.17$\\
\hline
\end{tabular}
\end{center}
\label{Tab_FDDB_Results}
}
\vspace*{-0.8cm}
\end{table*}

\begin{figure*}[!t]

\begin{center}
\includegraphics[width=0.9\linewidth]{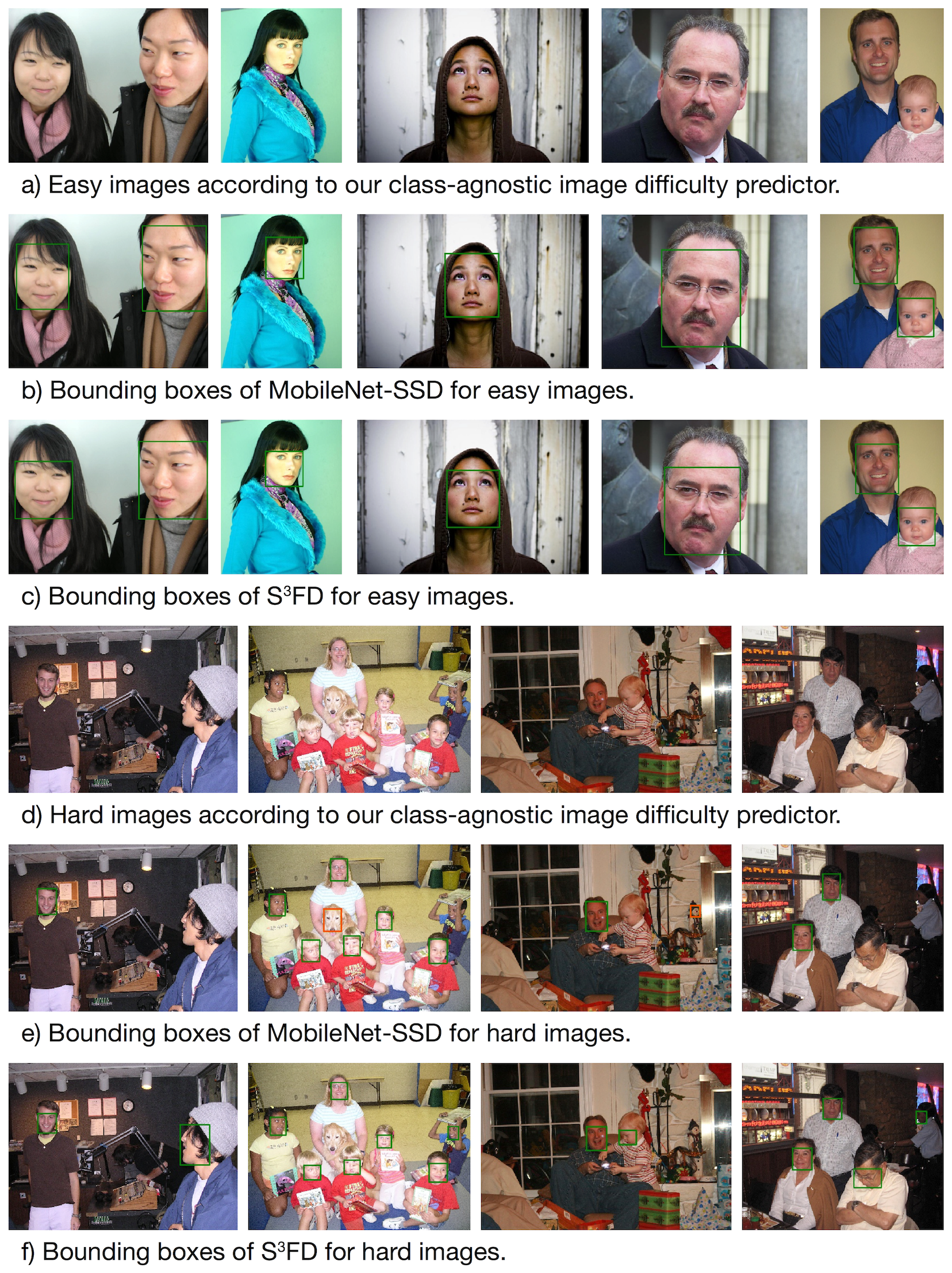}
\end{center}
\vspace*{-0.7cm}
\caption{Examples of easy (top three rows) and hard images (bottom three rows) from AFW according to the class-agnostic image difficulty. For each set of images, the bounding boxes predicted by the MobileNet-SSD~\cite{Howard-arXiv-2017} and the S$^3$FD~\cite{Zhang-ICCV-2017} detectors are also presented. The correctly predicted bounding boxes are shown in green, while the wrongly predicted bounding boxes are shown in red. Best viewed in color.}
\label{fig_easy_vs_hard}
\vspace*{-0.4cm}
\end{figure*}

Regarding our five easy-versus-hard strategies for combining face detectors, the empirical results indicate that the proposed splitting strategies give better performance than the random splitting strategy, on both data sets. Although using the number of faces or the average size of the faces as splitting criteria is better than using the random splitting strategy, it seems that combining the two measures into a single strategy ($n / avg$) gives better and more stable results. On the AFW data set, the $n / avg$ strategy gives the best results for the $75\%-25\%$ split ($0.9571$), while the class-agnostic image difficulty provides the best results for the $50\%-50\%$ split ($0.9818$) and the $25\%-75\%$ split ($0.9923$). The highest improvements over the random strategy can be observed for the $50\%-50\%$ split. Indeed, the results for the $50\%-50\%$ split shown in Table~\ref{Tab_AFW_Results} indicate that our strategy based on the class-agnostic image difficulty gives a performance boost of $4.63\%$ (from $0.9355$ to $0.9818$) over the random splitting strategy. Remarkably, the AP of the MobileNet-SSD and S$^3$FD $50\%-50\%$ combination is just $1.49\%$ under the AP of the standalone S$^3$FD, while the processing time is reduced by almost half. On the FDDB data set, the $n / avg$ strategy gives the best DiscROC score for the $75\%-25\%$ split ($0.9214$), while the class-agnostic image difficulty provides the best DiscROC score for the $25\%-75\%$ split ($0.9673$). The $n / avg$ strategy and the class-agnostic image difficulty provide equally good DiscROC scores on the $50\%-50\%$ split ($0.9493$). As indicated in Table ~\ref{Tab_FDDB_Results}, the best DiscROC score for the $50\%-50\%$ split ($0.9493$) on FDDB is $3.83\%$ above the DiscROC score ($0.9110$) of the baseline strategy. The ContROC scores are generally lower for all models, but the same patterns occur in the results presented in Table ~\ref{Tab_FDDB_Results}, i.e. the best results are provided either by the $n / avg$ strategy when MobileNet-SSD has a higher contribution in the combination or by the class-agnostic image difficulty when S$^3$FD has to process more images.

To understand why our splitting strategy based on the class-agnostic image difficulty scores gives better results than the random splitting strategy, we randomly select a few easy examples (with less and bigger faces) and a few difficult examples from the AFW data set, and we display them in Figure~\ref{fig_easy_vs_hard} along with the bounding boxes predicted by the MobileNet-SSD and the S$^3$FD models. On the easy images, both detectors are able to detect the faces without any false positive detections, and the bounding boxes of the two detectors are almost identical. Nevertheless, we can perceive a lot more differences between MobileNet-SSD and S$^3$FD on the hard images. In the left-most hard image, MobileNet-SSD is not able to detect the profile face of the man in the near right side of the image. In the second image, MobileNet-SSD wrongly detects the dog's face as a human face and it fails to detect the face of the boy sitting in the right. In the third image, MobileNet-SSD wrongly detects the small snowman's face sitting in the background and it fails to detect the face of the baby. In the right-most hard image, MobileNet-SSD fails to detect the face of the person looking down, which is difficult to detect because of the head pose. In the same image, MobileNet-SSD also fails to detect the profile face of the man in the far right side of the image. Remarkably, the S$^3$FD detector is able to correctly detect all faces in the hard images illustrated in Figure~\ref{fig_easy_vs_hard}, without any false positive detections. We thus conclude that the difference between MobileNet-SSD and S$^3$FD is only noticeable on the hard images. This could explain why our splitting strategy based on the class-agnostic image difficulty scores is effective in choosing an optimal trade-off between accuracy and speed.

\vspace*{-0.2cm}
\section{Conclusion}
\label{sec_Conclusion}
\vspace*{-0.3cm}

In this paper, we have presented five easy-versus-hard strategies to obtain an optimal trade-off between accuracy and speed in face detection from images. Our strategies are based on dispatching each test image either to a fast and less accurate face detector or to a slow and more accurate face detector, according to the class-agnostic image difficulty score, the person-aware image difficulty score, the number of faces contained in the image, the average size of the faces, or the number of faces divided by their average size. We have conducted experiments using state-of-the-art face detectors such as S$^3$FD~\cite{Zhang-ICCV-2017} or MobileNet-SSD~\cite{Howard-arXiv-2017} on the AFW~\cite{Zhu-CVPR-2012} and the FDDB~\cite{Jain-UMAS-2010} data sets. The empirical results indicate that using either one of the image difficulty predictors for splitting the test images compares favorably to a random split of the images. However, our other easy-versus-hard strategies also outperform the random split baseline. Since all the proposed splitting strategies are simple and easy to implement, they can be immediately adopted by anyone that needs a continuous accuracy versus speed trade-off optimization strategy in face detection. %In future work, we aim to investigate whether training face detectors to specifically deal with easy or hard image samples can help to further improve our results.

\vspace*{-0.3cm}
\subsubsection*{Acknowledgments.} %\vspace*{-0.2cm}
The work of Petru Soviany was supported through project grant PN-III-P2-2.1-PED-2016-1842. 
The work of Radu Tudor Ionescu was supported through project grant PN-III-P1-1.1-PD-2016-0787.

\vspace{-0,3cm}
\bibliography{references}{} 
\bibliographystyle{splncs03}

\end{document}